\documentclass[
]{ceurart}

\usepackage{lmodern}

\usepackage{todonotes}
\begin{document}

\copyrightyear{2021}
\copyrightclause{Copyright for this paper by its authors.
  Use permitted under Creative Commons License Attribution 4.0
  International (CC BY 4.0).}

\conference{CLEF 2021 -- Conference and Labs of the Evaluation Forum, 
	September 21--24, 2021, Bucharest, Romania}

\title{Transformer-based Language Models for Factoid Question Answering at BioASQ9b}




\author{Urvashi Khanna}[%
orcid=0000-0003-2345-5596,
email=Urvashi.Khanna@mq.edu.au,
]
\address{Macquarie University, Australia}

\author{Diego Moll\'a}[%
orcid=0000-0003-4973-0963,
email=Diego.Molla-Aliod@mq.edu.au,
url=https://researchers.mq.edu.au/en/persons/diego-molla-aliod,
]


\begin{abstract}
In this work, we describe our experiments and participating systems in the BioASQ Task 9b Phase B challenge of biomedical question answering. We have focused on finding the ideal answers and investigated multi-task fine-tuning and gradual unfreezing techniques on transformer-based language models. For factoid questions, our ALBERT-based systems ranked first in test batch 1 and fourth in test batch~2. Our DistilBERT systems outperformed the ALBERT variants in test batches 4 and 5 despite having 81\% fewer parameters than ALBERT. However, we observed that gradual unfreezing had no significant impact on the model's accuracy compared to standard fine-tuning.

\end{abstract}


\maketitle

\section{Introduction}

Nowadays, the use of language models that have been pretrained on massive amounts of data are the norm \cite{Devlin2018, lan2019albert, liu2019roberta}. Rather than making significant task-specific architecture improvements, these pretrained models can be fine-tuned for various tasks by making minor changes to the language model architecture, such as adding an output layer on top. Fine-tuning approaches are critical for learning the distributions of the target task and improving the language model's adaptability. However, fine-tuning a language model on small datasets like BioASQ can lead to catastrophic forgetting and overfitting. Furthermore, training all layers simultaneously on data of different target tasks may result in poor performance and an unstable model \cite{ruder2019transfer}. A schedule for updating the pretrained weights may be critical for preventing catastrophic forgetting of the source task's knowledge. Scheduling techniques like chain thaw \cite{felbo-etal-2017-using} and gradual unfreezing \cite{Howard2018} have improved the performance of multiple Natural Language Processing (NLP) tasks. Gradual unfreezing involves gradually fine-tuning model layers rather than fine-tuning all layers at once.

Pretrained language models are usually trained on general language and then adapted to downstream tasks of varied domains. Many domain-specific tasks, however, face the problem of the scarcity of labelled datasets. Auxiliary signal through multi-task fine-tuning helps the language model to adapt on smaller datasets better \cite{sun2019fine, garg2020tanda, kang2020transferability}. Multi-task fine-tuning (also referred to as sequential adaptation in some literature \cite{ruder2019transfer}) is the intermediate fine-tuning stage in which the model is fine-tuned on a larger dataset before fine-tuning on a low-resource dataset. In this paper, we describe the experiments of our participating systems\footnote{Code associated with this paper is available at \url{https://github.com/urvashikhanna/bioasq9b}} at the  BioASQ9b challenge\footnote{\url{http://bioasq.org/}}. We discuss two of our systems, mainly focusing on factoid questions. Both systems adapt the multi-task fine-tuning technique of fine-tuning on a larger dataset before fine-tuning on the BioASQ9b dataset. Our first system fine-tunes the pre-trained model ALBERT on SQuAD2.0 and then on the BioASQ9b dataset. This system performed exceedingly well on BioASQ9b Test batches 1 and 2. Our second system investigates the effect of the gradual unfreezing technique on the smaller, compact transformer-based model, DistilBERT.  We assess this system via two of our submissions at the BioASQ9b Challenge. One of our submissions of DistilBERT ranked sixth in the BioASQ9b leaderboard\footnote{\url{http://participants-area.bioasq.org/results/9b/phaseB/}}. From our results, we conclude that gradually unfreezing DistilBERT had no significant improvement in the accuracy of the BioASQ9b test data in comparison to standard fine-tuning.

The rest of this paper is structured as follows. In Section~\ref{related work}, we briefly discuss related work for background. Section~\ref{BioASQ} describes the BioASQ dataset and the processing steps involved. Section~\ref{systems} details our experimental setup for both our systems. Section~\ref{results} discusses the results of our systems on the BioASQ public leaderboard. Finally, Section~\ref{conclude} provides a conclusion to our work.

\section{Related Work}\label{related work}

Transfer learning has been widely used to transfer knowledge across multiple domains. The scarcity of sizable domain-specific datasets and the cost associated with manually annotating them are driving this trend. In this section, we discuss previous works that used transfer learning for the BioASQ biomedical question answering task \cite{Tsatsaronis2015}.

In the 5th BioASQ challenge, Wiese et al. \cite{wiese2017neural} explored domain adaptation to transfer knowledge from an already existing neural Question Answering (QA) system named FastQA \cite{weissenborn2017making} that was trained on SQuAD \cite{RajpurkarZLL16}. They initialised their model with the pretrained FastQA models' parameters during the fine-tuning phase. Using a combination of fine-tuning and biomedical Word2vec embeddings, their model achieved state-of-the-art results. They also used optimisation approaches such as L2 weight regularisation and forgetting cost term to minimise catastrophic forgetting.

Lee et al. \cite{lee2019biobert} discovered the potential to adapt the general domain language model BERT for the biomedical domain. They presented BioBERT, the first biomedical language model. In the pretraining step, BioBERT was initialised with BERT weights and then pretrained on biomedical domain corpora. BioBERT produced benchmark results on a wide range of biomedical text mining tasks, including question answering, relation extraction, and named entity recognition. Yoon et al.'s \cite{yoon2019pre} submission for task 7b topped the leaderboard in the 7th BioASQ challenge. They used a sequential adaptation technique in which pretrained BioBERT was fine-tuned first on the SQuAD dataset and then on the BioASQ dataset. 

Similarly, BioELMo \cite{jin2019probing} is a biomedical version of ELMo that outperforms BioBERT on the authors' probing tasks when used as a feature extractor. However, the fine-tuned BioBERT outperforms BioELMo on named entity recognition and Natural Language Inference (NLI) tasks.

Hosein et al. \cite{hosein2019measuring} studied domain portability and error propagation of BERT-based QA models through their BioASQ7b submissions. Their results concluded that general domain language models could generalise and give good results for domain-specific tasks. They also observed that pretraining is more critical than fine-tuning when improving the domain portability of BERT QA models. For yes/no questions in the BioASQ7 Phase B challenge, Resta et al. \cite{resta2019transformer} used an ensemble of classifiers with input from various transformer-based language models. They employed contextual embeddings from multiple pretrained language models, such as BERT and ELMO, as features to capture long-term dependencies.    

Jeong et al. \cite{kang2020transferability} expanded the prior work on BioBERT models \cite{lee2019biobert, yoon2019pre} in the 8th BioASQ challenge. They adapted multiple stages of fine-tuning by first fine-tuning BioBERT on the NLI dataset \cite{N18-1101}, then on the SQuAD dataset \cite{RajpurkarZLL16}, and finally on the downstream BioASQ dataset. Their results established that tasks like NLI that capture the relationships between sentence pairs improve the accuracy of the QA systems. Additionally, they analysed and reported the number of unanswerable questions from the BioASQ7b dataset in the QA setting. Kazaryan et al. \cite{kazaryantransformer} used ALBERT \cite{lan2019albert} as their base language model which was fine-tuned first on SQuAD v2.0 \cite{rajpurkar2018know}, and subsequently on the BioASQ8b data.

\section{BioASQ Data Processing}\label{BioASQ}

BioASQ \cite{Tsatsaronis2015} is an international biomedical challenge that comprises annual tasks on semantic indexing and biomedical question answering. The ninth BioASQ challenge consists of two shared tasks. Task 9a is a semantic indexing task that aims to annotate new PubMed articles automatically \cite{ pubmed} with Medical Subject Headings (MeSH). Task 9b is a question answering task devised for systems to answer four types of biomedical questions: factoid, summary, list, and yes/no. The participants are provided with questions along with relevant snippets. The output generated by their systems is either an exact answer (for yes/no, factoid, and list questions) or ideal answers (for summary questions), or both. The tasks are released in five batches over two months, with 24 hours to submit the answers after the release of each test batch.

\begin{figure}
\centering
\includegraphics[width=12cm]{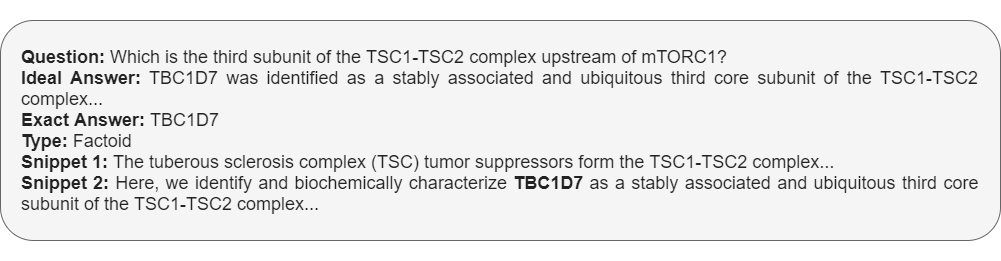}
\caption{Sample factoid question \cite{urvashi}. The answer to the question is in \textbf{bold} and is extracted from snippet 2.}
\label{figure1}
\end{figure}

We primarily concentrate on factoid questions from the BioASQ9b dataset. The dataset contains a total of 3743 questions, 1092 of which are factoid questions. An example of a factoid question is shown in Figure \ref{figure1}. Our system returns exact answers for factoid-type questions that can either be a single entity or a list of entities. We regard the BioASQ challenge task as an extractive QA task because the answer to the query is extracted from the relevant snippet. The metrics used for evaluating the systems on the BioASQ leaderboard are: Strict Accuracy (SAcc), Lenient Accuracy (LAcc), and Mean Reciprocal Rank (MRR). However, MRR is the official metric used by the BioASQ organisers for factoid questions since it is often used to evaluate other factoid QA tasks and challenges \cite{Tsatsaronis2015}. 

The BioASQ dataset is transformed into the SQuAD format and vice versa using pre-processing and post-processing steps. In a typical span-extractive question answering task, the system is provided with a passage P and a question Q, and it must identify an answer span A ($a_{start}, a_{end}$) in P. The SQuAD dataset is an example of a span prediction QA task containing many question-answer pairs and a passage that answers the given question. In contrast, the training dataset of BioASQ includes a question, an answer, and multiple relevant snippets. Therefore, we begin by pairing each snippet with its question and transforming it into multiple question-snippet pairs. Also, based on the exact answer provided, we locate the answer's position in the snippet and populate it as the start position of the answer span in the dataset. After performing these pre-processing steps, the BioASQ9b training data samples increased five-fold from 1092 to 5447. Table \ref{Table 3.1.1} shows the number of questions in the training and test batches before and after pre-processing.

\begin{table}[h]
\caption{Summary of BioASQ9b Training and Test data before and after pre-processing.}
\label{Table 3.1.1}
\centering
\begin{tabular}{lrr} 
\toprule
Dataset& \begin{tabular}[c]{@{}c@{}}Number of Factoid Questions\\Before Pre-processing\end{tabular} & \begin{tabular}[c]{@{}c@{}}Number of Factoid Questions\\After Pre-processing\end{tabular}  \\ 
\midrule
Training&1092&5447 \\
Batch 1&29&139 \\
Batch 2&34&151 \\
Batch 4&28&132 \\
Batch 5&36&148 \\
\bottomrule
\end{tabular}
\end{table}

Our system returns the prediction span for each question. Because we divided the snippets into several question-snippet pairs during the pre-processing stage, we now have predictions of multiple answer spans and their probabilities for each question. Each system must submit a list of up to five responses for the official BioASQ evaluation. As a result, we select the top five answers for each question in decreasing order of probability as our submission. Thus, for each factoid question, our system returns a list of up to five responses sorted by their likelihood.

\section{Systems Overview}\label{systems}

This section describes our systems and the experimental setup of our submissions at the BioASQ9b challenge. Our submissions in the BioASQ9b challenge are based on two pretrained models: ``DistilBERT'' and ``ALBERT''. As mentioned above, we focus mainly on factoid questions. We submitted ALBERT variants for all the BioASQ9b test batches except test batch 3. DistilBERT-based systems were submitted in test batches 2, 4, and 5. In this section, we detail the models, the methodology used, and the experimental setup.

\subsection{ALBERT}\label{albert}

\begin{figure}[h]
\centering
\includegraphics[width=14cm]{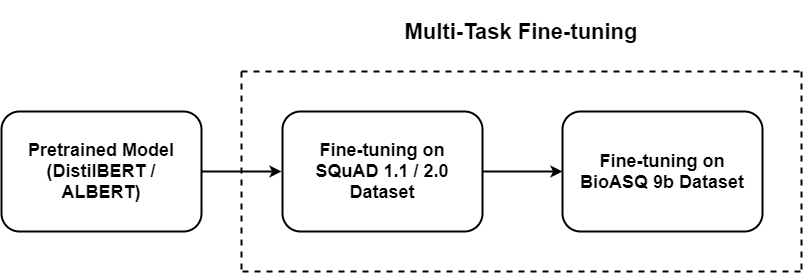}
\caption{Diagram depicting our system's fine-tuning strategy.}
\label{figure2}
\end{figure}

For the system using ALBERT, we follow a staged fine-tuning approach by fine-tuning on a large dataset before fine-tuning on the smaller dataset. This preliminary stage of fine-tuning on a large QA task is ideal due to the small size of the BioASQ dataset. However, large-scale bio-medical QA datasets are not readily available that could be used for the first stage of fine-tuning. Therefore, we use the SQuAD dataset, a widely used extractive QA dataset. Thus, we first fine-tune ALBERT on SQuAD2.0 and later on our downstream BioASQ task. This approach is illustrated in Figure \ref{figure2}.

ALBERT is a lighter version of BERT with considerably fewer parameters. Lan et al. \cite{lan2019albert} used two parameter-reduction strategies to lower the memory usage and increase the training speed of BERT. Since ALBERT models scale better than BERT, we have used the xxlarge version of ALBERT for our experiments. The BioASQ task was set up as a span-extraction QA task in which the model predicts the start and end span of answers for a given context and question. In both stages of fine-tuning, the input to the model is the concatenation of passage and question with a special token [SEP] separating them. This input is tokenized using WordPiece embeddings \cite{wu2016google} to handle the out-of-vocabulary issues. After WordPiece tokenization, the maximum allowable input sequence length is 512 for both the ALBERT and DistilBERT models. The input has three embeddings: token, position, and sentence. In order to differentiate between the sentences, sentence embedding is appended to each sentence, and a special position token is added to identify the position of each token. The model returns the start and end scores for each word. The output of the model is the candidate span with the highest score and where the end position is greater than or equal to the start position.

We employed ``ALBERT-xxlarge'' version 2 as our pretrained language model along with its tokenizer, which are publicly available from the Huggingface Transformers Library \cite{wolf-etal-2020-transformers}. This model has an additional task-specific linear question answering layer on top to output the start and end spans. Unless otherwise specified, the hyperparameters for both fine-tuning stages were set to the default values used by the ALBERT developers. The systems were validated on the BioASQ7b test batches 1 and 2.

All the three ALBERT-based submissions use the same fine-tuning approach discussed above with slight changes to the fine-tuning hyper-parameters. The systems along with hyperparameters are listed in Table \ref{Table 4.1} and their results are listed in Table \ref{Table 5.1}.

\begin{table}[h]
\caption{ALBERT-based systems along with the hyperparameters.}
\label{Table 4.1}
\centering
\begin{tabular}{lrrrr} 
\toprule
System Name & Learning Rate & Batch Size & Sequence Length & Epochs\\ 
\midrule
ALBERT 1&3e-5&4&512&3\\
ALBERT 2&2e-5&4&512&4\\
ALBERT 3&1e-5&4&512&3 \\
\bottomrule
\end{tabular}
\end{table}

\subsection{Gradual Unfreezing DistilBERT}\label{distilbert}

In recent years, the pretrained language models are getting bigger and deeper with millions, sometimes billions of parameters \cite{lan2019albert,radford2019language}. The success of these models on NLP tasks has fueled the race to scale up the models further. However, deploying these massive models on mobile and edge devices has implications such as environmental impact and computational cost  \cite{strubell2019energy}, making them unsuitable for use in real-world applications. Sanh et al. \cite{sanh2019distilbert} applied knowledge distillation \cite{hinton2015distilling} and proposed a smaller language model, DistilBERT, that achieves performance comparable to BERT on various NLP tasks.  DistilBERT, a distilled, compact version of BERT, has 60\% fewer parameters than BERT.


The focus for our second system was to study the effect of gradual unfreezing on the transformer-based language models. We used DistilBERT as our pretrained model to conduct the experiments of gradually unfreezing the transformer layers. The reason for this choice was the small size of DistilBERT and its ability to achieve close to 95\% of all the NLP task benchmarks when compared to BERT.

The process of fine-tuning allows the model to learn the distribution of the downstream task. In standard fine-tuning, all the layers of the model are trained on the target task simultaneously. Howard et al. \cite{Howard2018} introduced a fine-tuning approach of gradually unfreezing one layer at a time, starting from the top layer. They used a standard Long Short-Term Memory (LSTM) network without any attention mechanism for their experiments. Our work investigates the gradual unfreezing approach on DistilBERT using BioASQ9b as our target dataset.  

\begin{figure}[h]
\centering
\includegraphics[width=13cm]{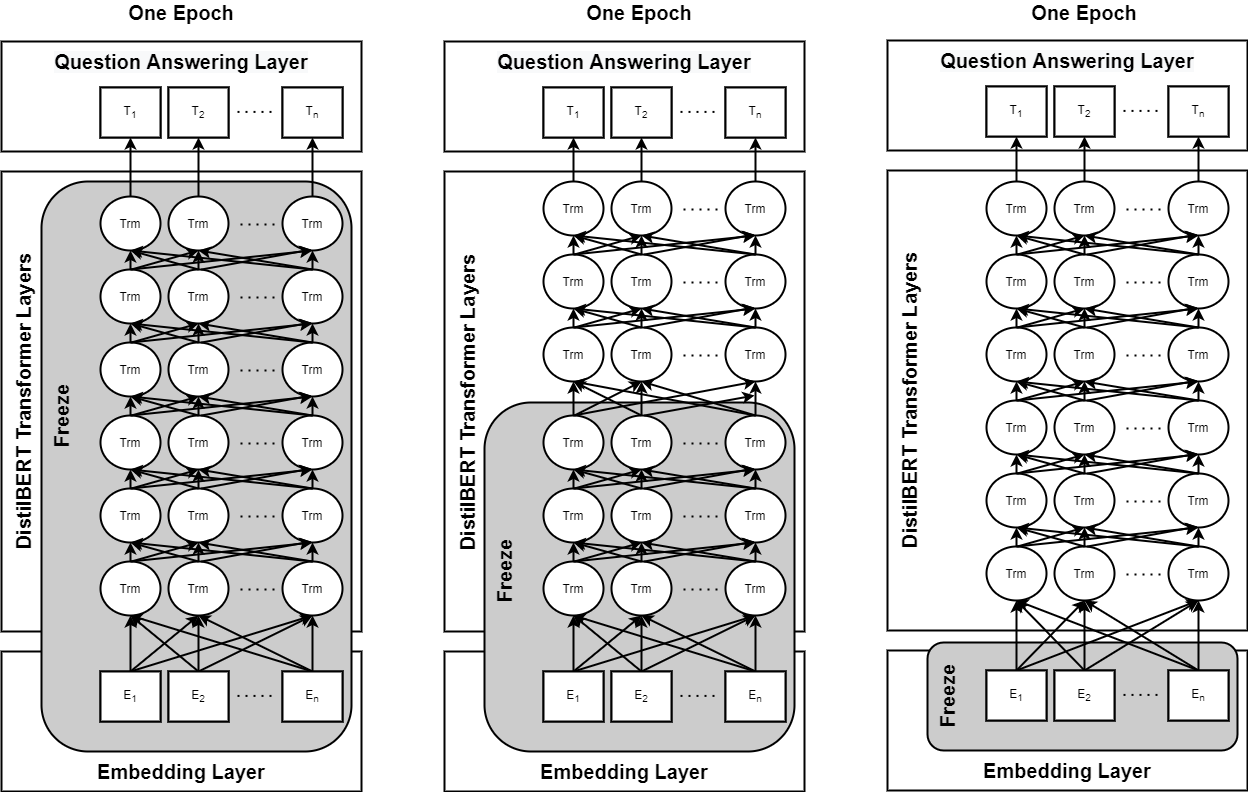}
\caption{Diagram showing our unfreezing approach.}
\label{Figure 3.3}
\end{figure}

DistilBERT has three blocks of layers: one embedding layer, six transformer layers, and a top task-specific layer. In our approach shown in Figure \ref{Figure 3.3}, we begin by fine-tuning only the top task-specific layer for one epoch while keeping all other layers frozen. Then we unfreeze the transformer layers consecutively in groups of three, fine-tune all the unfrozen layers for one epoch, and repeat until all layers are fine-tuned except the embedding layer. The decision to keep the embedding layer always frozen was based on the preliminary experiments in our previous work \cite{urvashi}. As a result, DistilBERT's trainable parameters have been reduced from 65 million to 42 million.   

In this system, ``distilbert-base-cased'' \cite{wolf-etal-2020-transformers} is first fine-tuned on SQuAD1.1 data and then on BioASQ9b task. Our gradual unfreezing approach is only applied during the second stage of fine-tuning. In the second phase of fine-tuning, we fine-tune the model at a constant learning rate of 3e-5, sequence length of 512, and for three epochs. We evaluate the unfreezing approach through two submissions at the BiOASQ challenge. The system ``DistilBERT'' is our baseline system. In this system, all the layers of DistilBERT are fine-tuned simultaneously. The system ``Unfreezing DistilBERT'' is the model that was fine-tuned using our unfreezing approach. Both systems are fine-tuned with the same hyperparameters for a fair comparison. Table \ref{Table 5.1} lists our systems with the results, along with the top-ranked system in the BioaASQ9b leaderboard. We have reported the MRR in the results table since it is the main metric used by the BioASQ organisers.

\section{Results}\label{results}

The results of our submissions to the BioASQ9b Phase B challenge are shown in Table \ref{Table 5.1}. From the results, we observe that ``ALBERT 2'' system was the best system for batch 1, and the ``ALBERT 3'' system was ranked fourth on the public leaderboard of the  BioASQ9b challenge. Overall, the systems using the pretrained ALBERT weights have performed exceedingly well on test batches 1 and 2. However, our ALBERT variants received poor results for test batches 4 and 5. It is worth noting that all the systems will be evaluated by humans experts after the competition. However, because this data was not accessible at the time of writing this study, we rely on automatic evaluations available on the BioASQ leaderboard.  

\begin{table}[h]
\caption{Results of our five submissions along with the top-ranked system from the BioASQ9b leaderboard. The first column of the table lists the unique submission identifier along with the system names as displayed on the public leaderboard. The highest score for each  batch is in \textbf{bold}.}
\label{Table 5.1}
\begin{tabular}{llllll}
\toprule
\multicolumn{1}{c}{\multirow{2}{*}{Submission ( Display name)}} & \multicolumn{1}{c}{\multirow{2}{*}{System}} & \multicolumn{4}{c}{Factoid - Mean Reciprocal Rank (MRR)} \\
\multicolumn{1}{c}{} & \multicolumn{1}{c}{} & Batch 1 & Batch 2 & Batch 4 & Batch 5 \\
\midrule
MQ TL1   (ALBERT) & ALBERT 1 & 0.4379 & 0.4667 & 0.369 & 0.4468 \\
MQ TL2   (Ensemble) & ALBERT 2 & \textbf{0.4632} & 0.501 & 0.4167 & 0.4731 \\
MQ TL-3   (Another ALBERT) & ALBERT 3 & 0.4621 & 0.5319 & 0.4375 & 0.4778 \\
MQ TL4 (Final   BERT) & DistilBERT & - & 0.5059 & 0.5399 & 0.5171 \\
MQ Transfer   Learning (MRes) & Unfreezing DistilBERT & - & 0.4887 & 0.5893 & 0.4917 \\
Top Ranked System  & - & \textbf{0.4632} & \textbf{0.5539} & \textbf{0.6929} & \textbf{0.588} \\
\bottomrule
\end{tabular}
\end{table}

The most noticeable difference between our DistilBERT and ALBERT variants, apart from their sizes, is the initial fine-tuning stage. In our systems, ALBERT was fine-tuned on SQuAD2.0, whereas DistilBERT was fine-tuned on SQuAD1.1. The SQuAD2.0 dataset is a reading comprehension dataset that, in addition to the SQuAD1.1 dataset, contains approximately 50,000 unanswerable questions. We need to look into whether test batches 1 and 2 had more unanswered questions after the organisers release the golden answers, and if so, how it has affected the results.

\begin{table}
\caption{Results from our previous work \cite{urvashi} on the BioASQ7b dataset. The system `KU DMIS Team' \cite{BioASQ7b_leaderboard, yoon2019pre} is BioBERT based system that was top of the leaderboard in the BioASQ7b challenge.}
\label{Table 5.2}
\centering
\begin{tabular}{lrr} 
\toprule
Systems & Mean Reciprocal Rank \\ 
\midrule
KU-DMIS Team \cite{BioASQ7b_leaderboard,yoon2019pre}  & 0.5235 \\
DistilBERT-fine-tuned & 0.4844 \\ 
DistilBERT-unfreeze-3 & 0.4841 \\
\bottomrule
\end{tabular}
\end{table}

From the results of Table \ref{Table 5.1}, we observe that both ``DistilBERT'' and ``Unfreezing DistilBERT''  outperformed the ALBERT variants for the test batches 4 and 5. Our system ``Unfreezing DistilBERT'' is ranked sixth in the BioASQ9b public leaderboard. The average MRR score of test batches 2, 4 and 5 for systems ``DistilBERT'' and ``Unfreezing DistilBERT'' is 0.5209 and 0.5232 respectively, and the difference is not statistically significant\footnote{Paired t-tests were used to compute the statistical significance since the MRR can be considered as a normal distribution as it is an average of samples. We find no statistically significant difference between the gradually unfrozen model and the baseline.}. Thus, we can conclude that gradually unfreezing the transformer-based models has no significant impact on the model's accuracy compared to typical fine-tuning. These results further support the findings of our previous work \cite{urvashi} on  gradually unfreezing DistilBERT with the BioASQ7b dataset, the results of which are shown in Table \ref{Table 5.2}. The results show that gradually unfrozen models produce promising results for a few test batches, but have no overall significant impact across all the test batches.

\section{Conclusion}\label{conclude}
Our participation in BioASQ9b was primarily focused on generating the ideal answers for factoid questions. We participated in four test batches, with our systems employing pretrained ALBERT and DistilBERT language models. The results were mixed, with ALBERT-based systems ranking amongst the top systems for test batches 1 and 2. For test batch 4, the compact DistilBERT variants, although having 81 percent fewer parameters, scored considerably better than ALBERT. This paves the way for a biomedical version of DistilBERT for mobile and edge devices for real life biomedical QA applications. In addition, we investigated the effect of gradual unfreezing on transformer-based language models using the BioASQ9b dataset. We conclude that gradually unfreezing the layers of DistilBERT had no significant impact on the model's accuracy in comparison to standard fine-tuning. We also investigated an unfreezing approach that makes use of only 66\% of DistilBERT's parameters when fine-tuning. In the future, we will aim to investigate ensemble or hybrid models of DistilBERT and ALBERT.

  

\bibliography{sample-ceur}

\appendix

\end{document}